\documentclass[preprint,review,12pt]{elsarticle}

\usepackage{amssymb}
\usepackage{amsmath}
\usepackage{amsthm}

\usepackage{bm}
\usepackage{algorithm}%
\usepackage{algorithmicx}%
\usepackage{algpseudocode}%
\usepackage[caption=false]{subfig}
\usepackage{threeparttable}
\usepackage{multirow}
\usepackage{hyperref}
\newtheorem{theorem}{Theorem}
\newtheorem{lemma}{Lemma}

\theoremstyle{definition}
\newtheorem{definition}{Definition}

\begin{document}

\begin{frontmatter}



\author[1]{Ke Chen} 
 \ead{kechen@stu.xjtu.edu.cn}

 \author[1]{Dandan Jiang\corref{cor1}}
 \ead{jiangdd@xjtu.edu.cn}
 \cortext[cor1]{Corresponding author.}
\affiliation[1]{organization={School of Mathematics and Statistics, Xi'an Jiaotong University},
	addressline={No.28, Xianning West Road}, 
	city={Xi'an},
	postcode={710049}, 
	state={Shaanxi},
	country={China}}

\title{Nonlinear Principal Component Analysis with Random  Bernoulli Features for Process Monitoring}


\begin{abstract}
The process generates substantial amounts of data with highly complex structures, leading to the development of numerous nonlinear statistical methods. However, most of these methods rely on computations involving large-scale dense kernel matrices. This dependence poses significant challenges in meeting the high computational demands and real-time responsiveness required by online monitoring systems.
To alleviate the computational burden of dense large-scale matrix multiplication, we incorporate the bootstrap sampling concept into random feature mapping and propose a novel random Bernoulli principal component analysis method to efficiently capture nonlinear patterns in the process.
We derive a convergence bound for the kernel matrix approximation constructed using random Bernoulli features, ensuring theoretical robustness. Subsequently, we design four fast process monitoring methods based on random Bernoulli principal component analysis to extend its nonlinear capabilities for handling diverse fault scenarios.
Finally, numerical experiments and real-world data analyses are conducted to evaluate the performance of the proposed methods. Results demonstrate that the proposed methods offer excellent scalability and reduced computational complexity, achieving substantial cost savings with minimal performance loss compared to traditional kernel-based approaches.
\end{abstract}




\end{frontmatter}



\section{Introduction}\label{Sec:Introduction}

The monitoring of the operation process of industrial plants and the fault detection in the process are the keys to maintaining the efficient and reliable operation of industrial plants \cite{8304817}. In recent years, the complexity and scale of industrial processes have increased dramatically. Through Supervisory Control and Data Acquisition systems, samples are taken every few seconds from online sensors with hundreds to thousands of process variables \cite{1035214}. 

The complex structure in the massive data generated in process monitoring has rendered traditional linear methods, such as principal component analysis (PCA), inadequate. This has driven the development of advanced statistical methods designed for nonlinear process monitoring.
To name a few, the {pioneering} work \cite{LEE2004223} applies the kernel PCA to nonlinear process monitoring, then kernel PCA has been widely used to deal with various aspects of nonlinear process monitoring \cite{GE20092245,YAO201556,Li2015EnsembleKP}.
As the complexity of the data increases, processes with temporal correlation and time-varying systems are gradually considered. For example,  
\cite{CHOI20045897,ZHANG2020107738} propose dynamic kernel PCA and introduce the time-lagged vector to extract the dynamic features in the nonlinear process.
To fit time-varying systems, \cite{Wang2005ProcessMA,JENG2010475} combine kernel PCA with a moving window to make the model adaptive.
All of the above kernel-based methods rely on all the normal operation condition samples to generate features, which results in a high computational cost for a kernel matrix of full sample selection.
For large-scale process monitoring, the real-time calculation of the large-scale matrix is even more difficult to load.

In order to improve the computational efficiency of these methods involving kernel matrices, a number of techniques for sample subset selection are proposed. \cite{191} uses projection to select transformed data that properly approximates the principal components of the kernel PCA model. \cite{246} proposes a new approximation criterion to select the suitable kernel functions, thereby reducing the number of kernel functions. \cite{249} performs a partially reduced kernel PCA model on a subset of variables to reduce computation time.
Although these methods speed up the calculation by reducing the number of samples, the approximate performance of the reduced kernel matrix is usually unstable, due to the randomness of sample subset selection. 
Another approach involves using low-rank approximations of large-scale kernel matrices to reduce the computational complexity of kernel PCA.
\cite{Williams2000UsingTN,IOSIFIDIS2016190} use the Nystr{\"o}m method to construct a low-rank matrix and approximate the original kernel matrix.
\cite{Rahimi2007RandomFF} maps the data to a low-dimensional space and approximates the kernel function by this mapping. Based on this, \cite{10.5555/3044805.3045044} uses random Fourier features to construct an approximate kernel matrix and extends it to multivariate statistical methods to propose random PCA.  {\cite{8649617} innovatively applies random PCA to online process monitoring to save computing costs. 
	However, the random Fourier features still rely on large-scale matrix multiplication, which is computationally costly and less helpful for real-time event detection.

	In this paper, we incorporate the bootstrap sampling concept into random feature mapping, which avoids the burden hard in dense large-scale matrix multiplication. A novel random Bernoulli PCA method is also proposed to capture nonlinear complex structure efficiently.
	To enable real-time responsiveness in online process monitoring with high accuracy and low computational cost, four fast process monitoring methods based on random Bernoulli PCA are designed for different fault scenarios:
	the random Bernoulli PCA for the static process monitoring; the dynamic random Bernoulli PCA and two-dimensional random Bernoulli PCA for dynamic process monitoring; the moving-window random Bernoulli PCA for a time-varying process monitoring. 
	The main contributions of our proposed work are as follows:	
	\begin{itemize}
		\item The random Bernoulli feature utilizes sparse matrix multiplication rather than dense matrix multiplication in random Fourier feature, significantly improving computational efficiency in large-scale online algorithms.
		Actually, the random Bernoulli feature applies the concept of randomly resampling from bootstrap to random feature mapping, offering broad scalability and enabling the use of many linear algorithms to solve nonlinear problems.
		\item We apply the random Bernoulli feature to PCA to obtain its nonlinear variant: random Bernoulli PCA.  
		An approximate kernel matrix is constructed using the random Bernoulli feature function. Furthermore, the spectral norm error between this approximation and the Gaussian kernel matrix is analyzed, ensuring that the approximate kernel matrix retains the advantages of the Gaussian kernel matrix. As a result, the random Bernoulli PCA achieves low computational cost while maintaining nearly the same performance superiority.	
		
		\item Compared with kernel PCA, the random Bernoulli PCA reduces computational complexity from $\mathcal{O}(n^3)$ to $\mathcal{O}(n)$, where $n$ is the sample size; Compared with random PCA, it replaces the dense Gaussian matrix with the sparse Bernoulli matrix, saving the complexity required for matrix multiplication. 
		The experimental results demonstrate that the proposed methods can achieve faster and more efficient monitoring than other kernel-based methods, and the modeling and online monitoring time are reduced by at least an order of magnitude.
	\end{itemize}
	
This paper is organized as follows:  
In Section \ref{Sec:Random Bernoulli PCA}, the random Bernoulli PCA is proposed for efficient extraction of nonlinear features, and the convergence error is analyzed to ensure sparsity without compromising performance.
In Section \ref{Sec:RBPCA-based}, the static, dynamic, and time-varying monitoring methods based on random Bernoulli PCA are proposed, respectively, and the  computational complexity of these methods are analyzed.
In Section \ref{Sec:Simu}, the monitoring performance is studied using numerical examples and the Tennessee Eastman process. The effectiveness is verified on a real data in Section \ref{Sec:Real data}. The conclusion is drawn  finally in the Section~\ref{sec:conc}.

\section{Nonlinear PCA Based on Random Bernoulli Feature}\label{Sec:Random Bernoulli PCA}
\subsection{Nonlinear Random Bernoulli Feature}
Random feature mapping is a technique that maps high-dimensional input data to a relatively low-dimensional random feature space, it is originally proposed by \cite{Rahimi2007RandomFF}. Combining random feature mapping with existing kernel algorithms can accelerate the training of large-scale kernel machines. Since then, a series of works \cite{pmlr-v22-kar12,pmlr-v32-hamid14} have been performed to approximate the kernel by using random feature mappings. 
To further reduce the computational complexity, we 
propose a random Bernoulli feature, which uses a sparse Bernoulli matrix to construct the feature.

Firstly, we consider the functional nonlinear random features, 
$f(\bm{X})=\int \alpha(\bm{B})\phi(\bm{X};\bm{B}) d\bm{B} $, where $\phi: \mathcal{X} \times \mathbb{R}^D \rightarrow \mathbb{R}$ is a nonlinear feature function, parameterized by some vector $\bm{B} \in \mathbb{R}^D$, and $\alpha: \mathbb{R}^D \rightarrow \mathbb{R}$ is a mapping of weights [see \cite{NIPS2008_0efe3284}]. More simply, we consider a finite number of scalar weights $\alpha$ and parameter vectors $\bm{B}$ form of $f$:
$f(\bm{X})=\sum_{j=1}^m \alpha_j\phi(\bm{X}^\top\bm{B}_j).$
The nonlinear feature function $\phi$ is chosen as the cosine function, and the parameter vectors $\bm{B}_j\ (j=1,\dots,m)$ are randomly sampled from the distribution $\mathbb{P}(\bm{B})$. For the data $\bm{X}=(\bm{x}_1,\dots,\bm{x}_n) \in \mathbb{R}^{D\times n}$,  the $m$ nonlinear random feature mappings $\widetilde{\bm{z}}_j(\bm{X})\ (j=1,\dots,m)$ are constructed by the following structure:
\begin{align}\label{zhat(x)}
	\widetilde{\bm{z}_j}(\bm{x}_k)=&\sqrt{2}\cos(\bm{x}_k^\top\bm{B}_j+u_j) \quad (k=1,\dots,n),
\end{align}
where $\widetilde{\bm{z}_j}(\bm{X})=(\widetilde{\bm{z}_j}(\bm{x}_1),\dots,\widetilde{\bm{z}_j}(\bm{x}_n))^\top$ and $u_j$'s are drawn uniformly from $(0,2\pi)$.

The random Fourier feature in \cite{Rahimi2007RandomFF} is the most well-known random feature mapping, where  $\mathbb{P}(\bm{B})$ is set as the Fourier transform of shift-invariant kernel. Some popular shift-invariant kernels are Gaussian kernel, Laplacian kernel and Cauchy kernel, among which the Gaussian kernel $k(\bm{x},\bm{y})=\exp(-\Vert \bm{x}-\bm{y}\Vert_2 ^2/c)$ is the most widely used kernel in the field of kernel machines, {with $c$ being} the kernel width parameter \cite{10.7551/mitpress/3206.001.0001}.
The random Fourier feature approximates the Gaussian kernel by setting $\mathbb{P}(\bm{B})$ to a Gaussian distribution $N(0,2\bm{I}/c)$,  involving $\mathcal{O}(nmD)$ complexity of matrix multiplication during the generation of this feature.
For accelerating the computation of $\bm{x}_k^\top\bm{B}_j$, we propose random Bernoulli feature by setting $\mathbb{P}(\bm{B})$ to Bernoulli distribution, namely { the components of $\bm{B}_j$ in (\ref{zhat(x)}) are independently} sampled from $b(1,p)$, where $p \in (0,1)$ is a probability parameter. 
Sparse matrix multiplication can reduce $(1-p)\%$ of the computational complexity compared with dense matrix multiplication. The later simulations show that these methods using random Bernoulli features can still achieve satisfactory performances when $p$ is as small as 0.05. Similar ideas can be found in bootstrap resampling and the construction of measurement matrix for compressed sensing.
Bootstrap estimates the distribution of the data population by randomly sampling the original dataset via multiple random Bernoulli vectors \cite{James2013AnIT}. 
\cite{10.5555/1756006.1859933,pmlr-v38-li15c} study compressed sensing based on a sparse random measurement matrix with a relatively small number of non-zero entries in each row, which can potentially reduce the complexity of signal recovery.

To preserve the invariant properties of the Gaussian kernel, which serve as a key theoretical foundation for asymptotic error analysis in process control,
we normalize the random Bernoulli vector $\bm{B}_j$, 
where the normalized random vector $(\bm{B}_j-p \bm{1}_{D \times 1})/\sqrt{cp(1-p)/2}$  asymptotically follows the Gaussian distribution $N(0,2\bm{I}/c)$.
Based on it, we give the definition of random Bernoulli feature as below.

\begin{definition}\label{approximate unbiased}
	The random Bernoulli feature is defined as follows: 	\begin{align}\label{zj(x)}
		\bm{z}_j(\bm{x}_k)=&\sqrt{2}\cos(\frac{\bm{x}_k^\top(\bm{B}_j-p\bm{1}_{D \times 1})}{\sqrt{cp(1-p)/2}}+u_j) \quad (k=1,\dots,n), 
	\end{align}
	where  $\bm{z}_j(\bm{X})=(\bm{z}_j(\bm{x}_1),\dots,\bm{z}_j(\bm{x}_n))^\top$ is formed of the matrix expression
	\begin{align}
		\bm{z}_j(\bm{X}) =&\sqrt{2}\cos(\frac{1}{\sqrt{cp(1-p)/2}}\bm{X}^\top\bm{B}_j-\frac{p}{\sqrt{cp(1-p)/2}}\bm{X}^\top\bm{1}_{D\times 1}+u_j\bm{1}_{n\times 1}).\nonumber
	\end{align}		
	Here $\bm{1}$ denotes a matrix with all elements 1, the components of $\bm{B}_j$ are independently sampled from a Bernoulli distribution $b(1,p)$, and $u_j$  is drawn uniformly from $(0,2\pi)$.  
\end{definition}

\subsection{Theoretical Justification}
Following the definition of random Bernoulli feature, it is obtained that  $\bm{z}_j(\bm{x}_k) \bm{z}_j(\bm{x}_l)$ is an approximate unbiased {estimator} of Gaussian kernel $k(\bm{x}_k,\bm{x}_l)$, and we also derive stronger convergence for every pair of points in the input space, more details of the proof are provided in supplementary material.
To further reduce the variance of the estimation, 
we construct $m$ random Bernoulli features $\bm{Z}(\bm{x}_k)=(\bm{z}_1(\bm{x}_k),\dots,\bm{z}_m(\bm{x}_k))$ and use the sample average $\bm{Z}(\bm{x}_k)\bm{Z}(\bm{x}_l)^\top/m$ to be an  approximation  of $k(\bm{x}_k,\bm{x}_l)$.
Let $\bm{K} \in \mathbb{R}^{n\times n}$ be the Gaussian kernel matrix of data $\bm{X}$,i.e.,$[\bm{K}]_{ij}=k(\bm{x}_i,\bm{x}_j)$, then the  kernel matrix $\bm{K}$ is  approximated by 
\begin{equation}\label{K_hat}
	\hat{\bm{K}}=\frac{1}{m}\sum_{j=1}^m \hat{\bm{K}}^{(j)}=\frac{1}{m}\sum_{j=1}^m \bm{z}_j(\bm{X}) \bm{z}_j(\bm{X})^\top.
\end{equation} 

To analyze the speed of the approximation $\hat{\bm{K}}$ converging to $\bm{K}$ in the spectral norm, we need prove 
the approximate matrix Bernstein's inequality first, which is formulated as a lemma with the detailed proof in supplementary material. 

\begin{lemma}[Approximate matrix Bernstein's inequality]\label{Bernstein}
	Consider an independent sequence $\bm{Y}_1,\dots,\bm{Y}_m$ of random matrices. Suppose that $\bm{Y}_k\in \mathbb{R}^{d\times d}\ (k=1,\dots,m)$ is a symmetric matrix satisfying that $E\left(\bm{Y}_k\right)=\bm{\epsilon}_k$,
	and existing a positive real number $R$ such that   
	$\Vert \lvert\bm{Y}_k\rvert \Vert \leq R$, where $\lvert\bm{Y}_k\rvert$ is the non-negative matrix with the absolute value of each element, and $\bm{\epsilon}_k$ is an error term and $\Vert \lvert \bm{\epsilon}_k\rvert \Vert\leq R/m$. Define the variance measure $\sigma^2=\Vert \sum_{k=1}^m E\left(\bm{Y}_k^2\right)\Vert$. Then, 
	\begin{footnotesize}
		$$E\left(\Vert\sum_{k=1}^m \bm{Y}_k\Vert\right) \leq \frac{R\log d}{\sqrt{2}} + \sqrt{2}R^2
		+\sqrt{\left(R+\log d \right)\left(3\sigma^2+2R^3+7R^2+\sqrt{2}R^2+\frac{7R^2}{m}+\frac{4R^2}{m^2}\right)} .$$
	\end{footnotesize}
\end{lemma}

Then the convergence rate  of the approximate error $\Vert \hat{\bm{K}}-\bm{K}\Vert$ is obtained based on the above lemma, and the detailed proof is provided in supplementary material.

\begin{theorem}\label{K_hat-K}
	Suppose that $[\bm{K}]_{ij}=k(\bm{x}_i,\bm{x}_j)$ is a kernel matrix constructed from a Gaussian kernel $k(\bm{x}_i,\bm{x}_j)=\exp(-\Vert \bm{x}_i-\bm{x}_j\Vert_2 ^2/c)$, and its approximation $\hat{\bm{K}}$ is formed by (\ref{K_hat}). 
	Then 
	\begin{small}
		$$\mathbb{E}\Vert \hat{\bm{K}}-\bm{K}\Vert  \leq \frac{\sqrt{2}n(m+1)\log n }{m(m-1)}+ \frac{4\sqrt{2}n^2(m+1)^2}{m(m-1)^2}+\sqrt{\frac{6n^2\log n}{m}+\frac{12n^3(m+1)}{m(m-1)}}.$$
	\end{small}
\end{theorem}
Since the solution of kernel PCA is the eigensystem of the kernel matrix $\bm{K}$, this theorem provides a theoretical guarantee for the subsequent application of random Bernoulli feature to PCA. 
Moreover, the computational complexity and spectral norm error of the approximate kernel matrices constructed by random Bernoulli feature and random Fourier feature are compared in supplementary material. The comparison results show that the error of the approximate kernel matrix constructed by random Bernoulli feature is almost the same as that of random Fourier feature, but the complexity saved is increase greatly with increasing $n$ or $m$.

\subsection{Random Bernoulli PCA}\label{RBPCA}
To achieve nonlinear dimensionality reduction of data, \cite{10.1007/BFb0020217,6790375} propose kernel PCA, which involves the high computational complexity of kernel matrix and is unsuitable for large-scale online monitoring applications.
To solve this problem, we propose random Bernoulli PCA based on random Bernoulli feature, which maps the nonlinear data into a lower dimensional feature space and then performs PCA on it. Specifically, we first randomly map the data $\bm{X}$ to a feature space $\mathcal{Z}$ through $\bm{z}_j(\bm{X})\ (j=1,\dots,m)$ in (\ref{zj(x)}). Let $\bm{Z}(\bm{X})=(\bm{z}_1(\bm{X}),\dots,\bm{z}_m(\bm{X}))$, then
$\bm{Z}: \mathbb{R}^{D\times n} \rightarrow \mathbb{R}^{n\times m}.$
Then PCA is performed in space $\mathcal{Z}$, that is, 
\begin{equation}\label{RBPCAeq}
	\text{random-Bernoulli-PCA}(\bm{X})=\text{PCA}\{\bm{Z}(\bm{X})\}.
\end{equation}
The mean centering procedure in the feature space $\mathcal{Z}$ should be performed, in which 
the vector $\bm{Z}(\bm{x}_k)$  is substituted by
\begin{equation}\label{Z_tilde}
	\overline{\bm{Z}}(\bm{x}_k)=\bm{Z}(\bm{x}_k)-\frac{1}{n}\sum_{j=1}^n\bm{Z}(\bm{x}_j) \quad (k=1,\dots,n).
\end{equation}
In the following text, $\overline{\bm{Z}}$ represents mean-centered $\bm{Z}$. Define $m \times m$ covariance matrix $\bm{R}$ by $\bm{R}=\overline{\bm{Z}}(\bm{X})^\top\overline{\bm{Z}}(\bm{X})/(n-1)$. The calculation of principal components is reduced to an eigenvalue problem: $\bm{R} \bm{v}=\lambda \bm{v}$. The $j$th principal component is calculated by $\overline{\bm{Z}}(\bm{X})\bm{v}_j $, where $\bm{v}_j$ is the eigenvector corresponding to the $j$th largest eigenvalue $\lambda_j$ of $\bm{R}$. Since the random Bernoulli feature used in random Bernoulli PCA only involves sparse matrix multiplication and random Bernoulli PCA is linear complexity in sample size, it provides the possibility to realize efficient large-scale online monitoring.

\section{Nonlinear Process Monitoring }\label{Sec:RBPCA-based}
\subsection{Static Process Monitoring}\label{Sec:RBPCA-based static} 
Fault detection is an important and indispensable key for the early identification of anomalies in process monitoring. We will implement the framework of fault detection with random Bernoulli PCA in this section.
We adopt the industrial datasets used for model training, which typically consist of samples collected under normal operation conditions: $\bm{x}_k\in \mathbb{R}^{D\times 1}\ (k=1,\dots,n)$.
Then we use random Bernoulli PCA in (\ref{RBPCAeq}) to transform the data into a reduced set of mutually independent features, namely the first $a$ nonlinear components are denoted by
$(\bm{t}_1,\dots,\bm{t}_a)=\overline{\bm{Z}}(\bm{X})\bm{V},$
where $\bm{V}=(\bm{v}_1,\dots,\bm{v}_a)$ {is consist of the corresponding $a$ eigenvectors} of $\bm{R}$.
The $a$-dimensional space spanned by $\bm{V}$ is called principal component subspace and the residual subspace is spanned by the remaining {vectors} $(\bm{v}_{a+1},\dots,\bm{v}_m)$.
The mapped data matrix $\overline{\bm{Z}}(\bm{X})$ can be projected into these two subspaces by random Bernoulli PCA, i.e.
$\overline{\bm{Z}}(\bm{X})=\widehat{\bm{Z}}(\bm{X})+\bm{E},$
where $\widehat{\bm{Z}}(\bm{X})=\sum_{j=1}^a \bm{t}_j\bm{v}_j^\top$ is the estimation of the mapped data matrix on the principal component subspace, and $\bm{E}$ is the projection of the mapped data matrix on the residual subspace.
Therefore, for the single sample $\bm{x}_k$, the estimation of the mapped data vector $\overline{\bm{Z}}(\bm{x}_k)$ is expressed as $\widehat{\bm{Z}}(\bm{x}_k)=\bm{V}\bm{V}^\top\overline{\bm{Z}}(\bm{x}_k)$.

The extracted nonlinear components $(\bm{t}_1,\dots,\bm{t}_a)$ are most sensitive to process faults, so the fault detection indicator based on these nonlinear components can be established referred to \cite{pr8010024}.
The commonly used $Q$ statistic, also known as the squared prediction error, indicates how well each mapped sample $\bm{Z}(\bm{x}_k)$ conforms to the random Bernoulli PCA model, i.e.
\begin{equation}\label{Q}
	Q_k=||\overline{\bm{Z}}(\bm{x}_k)-\widehat{\bm{Z}}(\bm{x}_k)||_2^2=\overline{\bm{Z}}(\bm{x}_k)^\top(\bm{I}-\bm{V}\bm{V}^\top)\overline{\bm{Z}}(\bm{x}_k),
\end{equation}
where $\bm{I}$ is an $m$-dimensional unit matrix.
It determines whether a process is faulty by comparing the gap between the new observation data $\bm{x}_{new}$ and the normal operation condition samples.
Therefore, the detection threshold, also known as the upper control limit, can be determined from a large number of normal operation condition samples. That is, the upper $\alpha$ quantile of the distribution of their $Q$ values is selected as upper control limit.
Every time a new sample $\bm{x}_{new}$ is collected, the value of its statistic is calculated through (\ref{Q}). When the calculated  $Q_{new}$ value exceeds the detection threshold, an alarm will be triggered, indicating the presence of a fault. 

In summary, the static process monitoring based on random Bernoulli PCA consists of two stages: Firstly, the modeling stage involves calculating the detection threshold using normal operation condition samples (see Algorithm \ref{alg: NOC model}); Secondly, the online monitoring stage assesses new data sampled at each time to determine whether there is a fault (see Algorithm \ref{alg: Online monitoring}). The selection of parameter $c$ in random Bernoulli feature refers to \cite{pr8010024}, see supplementary material for details.

\begin{algorithm}[htb]
	\caption{The modeling stage of static process monitoring based on random Bernoulli PCA.}\label{alg: NOC model}
	\begin{algorithmic}[1]
		\Require Normal operation conditions samples $\bm{X}$, number of random Bernoulli feature $m$, and significance level $\alpha$
		\Ensure Detection threshold $Q_{UCL}$
		\State Normalize data matrix $\bm{X}$;
		\State Randomly generate Bernoulli variables $\bm{B}_1,\dots,\bm{B}_m$ and uniform variables $u_1,\dots,u_m$;
		\State Calculate the random Bernoulli features $\bm{Z}(\bm{X})=(\bm{z}_1(\bm{X}),\dots,\bm{z}_m(\bm{X}))$ in (\ref{zj(x)});	
		\State Calculate the covariance matrix $\bm{R}=\overline{\bm{Z}}(\bm{X})^\top\overline{\bm{Z}}(\bm{X})/(n-1)$ and perform the eigenvalue decomposition;
		\For{$k\leftarrow 1$ to $n$}
		\State calculate $Q_k$ in  (\ref{Q});
		\EndFor
		\State Estimate the probability density functions of  $(Q_1,\dots,Q_n)$, find the upper $\alpha$ quantile $Q_{UCL}$.
	\end{algorithmic}	
\end{algorithm}

\begin{algorithm}[htb]
	\caption{The online monitoring stage of static process monitoring based on random Bernoulli PCA.}\label{alg: Online monitoring}
	\begin{algorithmic}[1]
		\Require New samples $\bm{x}_{new}$ and detection threshold $Q_{UCL}$
		\Ensure The process is faulty or normal
		\State Normalize the new data $\bm{x}_{new}$ with the	mean and variance obtained at step 1 of Algorithm \ref{alg: NOC model};
		\State Calculate the random Bernoulli features $\bm{Z}(\bm{x}_{new})=\left(\bm{z}_1(\bm{x}_{new}),\dots,\bm{z}_m(\bm{x}_{new})\right)$ in (\ref{zj(x)}) using the random variables generated in step 2 of Algorithm \ref{alg: NOC model};
		\State Obtain the mean-centered feature= $\overline{\bm{Z}}(\bm{x}_{new})=\bm{Z}(\bm{x}_{new})-\bm{Z}(\bm{X})^\top\bm{1}_{n\times 1}/n$, where $\bm{Z}(\bm{X})$ comes from the step 3 of Algorithm \ref{alg: NOC model};
		\State Calculate $Q_{new}$ in (\ref{Q})and monitor whether $Q_{new}$ exceeds its detection threshold $Q_{UCL}$.		
	\end{algorithmic}
\end{algorithm}

\subsection{Dynamic Process Monitoring}\label{Sec:RBPCA-based dynamic}
In industrial processes, variables are affected by random noise and uncontrollable perturbations, and exhibit some degree of autocorrelation.
The above static monitoring method only uses the measured value at the current time to evaluate the process each time, which does not consider the relationship between the measured values at different time points. So it cannot be effectively applied to the situation of dynamic process monitoring.
To improve the accuracy of fault detection for dynamic processes, we apply random Bernoulli PCA to the time-lagged vector data to extract the time-dependent relationship, making it adapt to the dynamic situation.

The past and current values of the measured variables are augmented as a time-lagged vector
\begin{equation}\label{time-lagged vector}
	\bm{y}_t=\left(\bm{x}_{t-l}^\top,\dots,\bm{x}_t^\top\right)^\top \in \mathbb{R}^{D(l+1)},
\end{equation}
where $l$ is the time lag. Dynamic process variables have temporal correlation, they are correlated in a certain time interval with relationship: $\bm{d}^\top\bm{Z}(\bm{y}_t)=0 $, where $\bm{d}$ is the parameter vector.
The dataset after the time-lagged vectors replace the original samples can be represented as
\begin{equation}
	\label{matrix Y}
	\bm{Y}=\left(\bm{y}_{l+1},\dots,\bm{y}_n\right) \in \mathbb{R}^{D(l+1) \times (n-l)},
\end{equation}
each column vector in the time-lagged dataset $\bm{Y}$ contains samples from different sampling times.
The method based on dynamic random Bernoulli PCA uses $\bm{Y}$ instead of $\bm{X}$ as the input data matrix, which is beneficial to describe the dynamic properties of the process.
In the online stage, every time a new sample $\bm{x}_{n+1}$ is obtained, the time-lagged vector $\bm{y}_{n+1}$ is constructed as the object of monitoring. Space is limited, and the specific algorithm is in supplementary material.

Dynamic random Bernoulli PCA extracts the dynamic properties in the process through the extended time-lagged vector and performs better when the time lag is small.
However, when the time lag is large, using PCA based on matrix-to-vector conversion results in a higher-dimensional vector space and a significant loss of structural information.
Therefore, we refer to two-dimensional PCA in \cite{1261097} and propose a process monitoring method based on two-dimensional random Bernoulli PCA as follows.

The random Bernoulli features $\bm{Z}(\bm{X})=\left(\bm{Z}(\bm{x}_1),\dots,\bm{Z}(\bm{x}_n)\right)^\top$ are constructed, then the past and current observations are formed into a time-lagged matrix instead of a vector, i.e.
\begin{equation}\label{time-lagged matrix}
	\bm{A}_t=\left(\bm{Z}(\bm{x}_{t-l}),\dots,\bm{Z}(\bm{x}_t)\right)^{\top}\in \mathbb{R}^{(l+1)\times m}. 
\end{equation}
After mean centering procedure, we define the following matrix 
\begin{equation}\label{2D covariance}
	\bm{G}=\frac{1}{n-l}\sum_{j=l+1}^n \overline{\bm{A}}_j^\top \overline{\bm{A}}_j
\end{equation}
similar to the image covariance matrix in \cite{1261097}.
By eigenvalue decomposition of $\bm{G}$, the obtained eigenvalues are arranged in descending order as $\sigma_1,\dots,\sigma_m$, and the corresponding eigenvectors are $\bm{p}_1,\dots,\bm{p}_m$.
After determining the number of principal components $a(a<m)$,  the principal component matrix $\bm{C}_t=\overline{\bm{A}}_t\bm{P}$ is obtained, where $\bm{P}=(\bm{p}_1,\dots,\bm{p}_a)$.
The estimation of $\overline{\bm{A}}_t$ on the principal component subspace spanned by $\bm{P}$ is $\widehat{\bm{A}}_t=\overline{\bm{A}}_t \bm{P} \bm{P}^\top$.

Similar to the Q statistic defined in (\ref{Q}), we define the two-dimensional $Q$ statistic as follows, which is a measure in each time-lagged matrix $\overline{\bm{A}}_t$ not captured by the principal components matrix retained $\bm{C}_t$, i.e
\begin{equation}\label{2D Q}
	Q^{2D}(t)=\text{tr}\left\{(\overline{\bm{A}}_t-\widehat{\bm{A}}_t)(\overline{\bm{A}}_t-\widehat{\bm{A}}_t)^\top\right\}=\text{tr}\left\{\overline{\bm{A}}_t (\bm{I}_m-\bm{P} \bm{P}^\top) \overline{\bm{A}}_t^\top \right\}.
\end{equation}
We calculate the two-dimensional $Q$ statistic value for normal operation condition samples and use kernel density estimation to determine the upper control limit of fault detection. The above is the dynamic process monitoring based on two-dimensional random Bernoulli PCA, and the algorithm is summarized in supplementary material. 

\subsection{Time-Varying Process Monitoring}\label{Sec:RBPCA-based time-varying}
The process behavior in practice is constantly changing over time, and the above monitoring methods are not applicable because the projection matrix and upper control limit are unchanged. Therefore, combined with moving windows, we formulate a scheme: moving-window random Bernoulli PCA, which makes the model adaptive to the time-varying process. This method is also divided into two stages: modeling and online monitoring.

In the modeling stage, the dataset is screened according to the similarity measurements to save subsequent computation. The cosine of the angle between the two vectors is used to measure the similarity between the input data
\begin{equation}\label{cosine}
	\cos(\bm{x}_i,\bm{x}_j)=\frac{\bm{x}_i \cdot \bm{x}_j}{\Vert \bm{x}_i\Vert_2 \Vert \bm{x}_j\Vert_2} \quad (i=1,\dots,n;\ j=1,\dots,n),
\end{equation}
where $\cdot$ denotes the dot product. Choosing a window width $w$, we keep the $w$ most dissimilar data and get the screened dataset $\bm{X}_s=\left\{ \bm{x}_{s_1},\dots, \bm{x}_{s_w} \right\}$.
Then the random Bernoulli PCA monitoring model is constructed on this screened dataset.

In the online monitoring stage, when the process produces a new observation $\bm{x}_{n+1}$, the detection indicator $Q_{n+1}$ can be calculated by (\ref{Q}). If $\bm{x}_{n+1}$ is the fault, the monitoring continues to the next sample $\bm{x}_{n+2}$.
If the process is normal at this time, we determine whether the model needs to be updated based on the following {criterion}. Firstly, we calculate the projection of $\bm{Z}(\bm{x}_{n+1})$ onto the space spanned by $\bm{Z}(\bm{X}_s)=\left\{\bm{Z}\left(\bm{x}_{s_1}\right),\dots,\bm{Z}\left(\bm{x}_{s_w}\right)\right\}$, denoted
\begin{equation}\label{projection}
	\widehat{\bm{Z}(\bm{x}_{n+1})}=\left(\bm{Z}(\bm{x}_{n+1}) \cdot \bm{Z}\left(\bm{x}_{s_1}\right),\dots,\bm{Z}(\bm{x}_{n+1}) \cdot \bm{Z}\left(\bm{x}_{s_w}\right) \right).
\end{equation}
If $\widehat{\bm{Z}(\bm{x}_{n+1})}$ does not approximate $\bm{Z}(\bm{x}_{n+1})$ well, it does not satisfy condition 
\begin{equation}\label{condition}
	\lvert \Vert\widehat{\bm{Z}(\bm{x}_{n+1})}\Vert_2-\Vert\bm{Z}(\bm{x}_{n+1})\Vert_2 \rvert <\delta,
\end{equation}
where $\delta$ is a given threshold. It means the new observation contains new relevant information about the process, and the screened dataset needs to be updated.
The updated dataset is $\bm{X}_{n+1}=\left\{\bm{x}_{s_2},\dots, \bm{x}_{s_w},\bm{x}_{n+1}\right\}$ by adding the latest observation and removing the oldest observation. 
Next, the random Bernoulli PCA monitoring model is established on this updated dataset to get the updated projection matrix and the updated upper control limit. The algorithm details steps are in given supplementary material.

\subsection{Computational Complexity Analysis}
\label{Sec:complexity}
The generation of random Bernoulli features mainly depends on a sparse Bernoulli matrix, so the computational complexity of this step is $\mathcal{O}(nmDp)$. The random Fourier features are generated using a dense Gaussian matrix with a complexity of $\mathcal{O}(nmD)$. In comparison, random Bernoulli features save $\mathcal{O}(nmD(1-p))$ computations,
but  retain strong feature extraction capability as illustrated in the simulations,
where  $p$ can be chosen very small, even as low as 0.05.

We analyze the overall computational complexity of random Bernoulli PCA as $\mathcal{O}(nmDp)+\mathcal{O}(m^2n)+\mathcal{O}(m^3).$
Similarly, the complexity of random PCA can be calculated as $\mathcal{O}(nmD)+\mathcal{O}(m^2n)+\mathcal{O}(m^3)$. The complexity of kernel PCA can be roughly calculated as $\mathcal{O}(n^2D)+\mathcal{O}(n^3)$. In general, the number of features $m$ is chosen to be much smaller than the number of samples $n$, so random Bernoulli PCA and random PCA are both linear in the sample size $n$, while kernel PCA remains the cubic in sample size.
In addition, for each new sample in the online monitoring stage, the method based on random Bernoulli PCA requires $\mathcal{O}(mDp)$ computational complexity to extract random Bernoulli features. The method based on random PCA requires $\mathcal{O}(mD)$ and the method based on kernel PCA requires $\mathcal{O}(nD)$.
Those extension methods have the same level of complexity as monitoring methods based on random Bernoulli PCA.

Therefore, constructing random Bernoulli features saves $(1-p)\%$ of complexity compared to random Fourier features. We initially map the data to a relatively low-dimensional feature space via random Bernoulli features and then apply the associated linear techniques in this space.
This enables us to address nonlinear problems while significantly improving computational speed, which scales linearly with the sample size $n$.

\section{Simulation Studies}
\label{Sec:Simu}
\subsection{Datasets and Performance Metrics}
We describe the two datasets used in the experiment and the performance metrics used to assess the results.
The first dataset is the numerical example as the following system with three variables originally proposed by \cite{752266}: $(x_1, x_2, x_3)^\top=(t, t^2-3t, -t^3+3t^2)^\top+(e_1, e_2, e_3)^\top$,
where $e_j\sim N(0,0.01)\ (j=1,2,3)$ are independent noise variables and $t$ is uniformly sampled from $[0.01,2]$.
Under normal operating conditions, 1000 samples are generated from this process as training data.
In addition, two sets of test data containing 500 samples are generated, with the following two faults, respectively.

Fault 1: Change the step size of $x_1$ by -0.5 starting from sample 201.

Fault 2: Add $0.01(j-200)$ to variable $x_2$ starting from sample 201, where $j$ is the sample number.

The second dataset is the Tennessee Eastman Process, developed by Eastman Chemical Company that simulates real chemical processes. This is the simulation case study most commonly used to test fault monitoring for complex industrial processes.
There are 11 manipulated variables and 41 process measurements, for a total of 52 variables.
The training and test samples are obtained under the 48-hour running simulation, with a sampling interval of 3 minutes, and 960 observations are collected. There are 21 types of fault scenarios, fault occurred at the $8$th hour of the test sample, namely the fault sample is after the $160$th sample.

To evaluate the performance of the process monitoring methods, we used these two metrics: 
Fault detection rate is the percentage of fault samples that are correctly detected as faults;
False alarm rate is the percentage of normal samples that are incorrectly detected as faults.
To measure the computational complexity and efficiency of each method, the time of normal operation condition modeling and the average time of online sample processing are compared.
For each performance metric, 500 Monte Carlo simulations are performed.

\subsection{The Performance of Process Monitoring}
In this subsection, we use the above datasets to verify the validity of the proposed monitoring algorithms based on random Bernoulli PCA. All simulations are run under Windows 10 and MATLAB R2020b. We compare the proposed four methods based on random Bernoulli feature with the other three kernel-based process monitoring methods: kernel PCA \cite{LEE2004223}, dynamic kernel PCA \cite{CHOI20045897}, moving-window reduced kernel PCA \cite{191}, and a method based on random Fourier feature: random PCA \cite{8649617}.
The number of principal components is selected according to the cut-off method in \cite{LEE2004223}. 
The radial basis function is used as the kernel function in kernel-based methods. The significance level $\alpha$ is set to 99\%. With little loss of performance, the parameter $p=0.05$ of the Bernoulli distribution and the number $m=150$ of random features are chosen for lower computational complexity. 
The time lag is set to $l=2$ (dynamic kernel PCA and dynamic random Bernoulli PCA) and $l=10$ (two-dimensional random Bernoulli PCA) in the numerical example, $l=8$ in the Tennessee Eastman Process. The window width is set to $w=500$ and $\beta=0.8$ for moving-window random Bernoulli PCA.

Table \ref{tab: Fault1_FDR} and Table \ref{tab:TEP} show the monitoring results of various methods on the numerical example dataset and the Tennessee Eastman Process dataset, respectively.
The methods proposed in this paper and the most accurate values on each dataset are bolded for ease of comparison. "Accurate"  here means that the fault detection rate is maximized under the premise of effectively inhibiting the false alarm rate (false alarm rate is less than 0.05). Table \ref{tab: time} shows the run time spent in the modeling stage and the online monitoring stage for each process monitoring method.

\begin{table*}
	\caption{The accuracy of online monitoring, including fault detection rate (FDR) and false alarm rate (FAR), is compared on the numerical example.}
	\renewcommand\arraystretch{1}
	{	\resizebox{\linewidth}{!}{
			\begin{threeparttable}
				\begin{tabular}{ccccccccccccc}
					\hline
					\multirow{2}{*}{Fault}&&  \multicolumn{2}{c}{KPCA}  &  \multicolumn{2}{c}{RPCA}  &  \textbf{RBPCA}  & \multicolumn{2}{c}{DKPCA}  & \textbf{DRBPCA} & 
					\textbf{2D-RBPCA} & MV-RKPCA& \textbf{MV-RBPCA} \\
					& & $T^2$& $Q$&$T^2$ &$Q$ &$Q$ &$T^2$ &$Q$ &$Q$ & $Q$& $Q$ & $Q$\\
					\hline
					\multirow{2}{*}{Fault 1}&FDR&  0.8069&  0.6248&  0.6019&   0.7297&   0.8917& 0.7159& 0.7861 &  \textbf{0.9566}&  0.9528& 0.6700&0.8367\\
					&FAR&  0.0113&  0.0114&  0.0103&  0.0115&    0.0117&  0.0113&  0.0129&  0.0126&  0.0165& 
					0.0250&0.0120\\
					\hline
					\multirow{2}{*}{Fault 2}& FDR&  0.8485&  0.7459&  0.6327&  0.8634&   0.8428&  0.8386& 0.8280& 0.8511 & \textbf{0.8646}& 0.7133&0.8311\\
					& FAR& 0.0104& 0.0104& 0.0100& 0.0115& 0.0109& 0.0109& 0.0134& 0.0118 & 0.0131& 0.0150
					&0.0073\\
					\hline
				\end{tabular}
				\label{tab: Fault1_FDR}
				\begin{tablenotes}
					\item	The proposed methods: RBPCA, random Bernoulli PCA; DRBPCA, dynamic random Bernoulli PCA; 2D-RBPCA, two-dimensional random Bernoulli PCA; MV-RBPCA, moving-window random Bernoulli PCA, all of which have been bold.
					Other methods: KPCA, kernel PCA; RPCA, random PCA; DKPCA, dynamic kernel PCA; MV-RKPCA, moving-window reduced kernel PCA. The most accurate values in the monitoring results for each fault have been bolded. The abbreviation is also applicable to other tables and figures.
				\end{tablenotes}
	\end{threeparttable}}}
\end{table*}

The results show that when the process exhibits simple dynamics, as in the numerical example (see Table \ref{tab: Fault1_FDR}), these time-invariant methods can all detect simulated faults at roughly the same level of accuracy. The dynamic PCA methods can observe more correlations, so the ability of dynamic random Bernoulli PCA and two-dimensional random Bernoulli PCA methods to detect faults is much higher than other methods. Time-varying methods are slightly "behind" their time-invariant counterparts because of the slightly older local realizations of the process, resulting in slightly worse performance.

\begin{table*}
	\caption{The accuracy of online monitoring, including fault detection rate (FDR) and false alarm rate (FAR), is compared on the Tennessee Eastman Process.}
	\renewcommand\arraystretch{1}
	{	\resizebox{\linewidth}{!}{
			\begin{threeparttable}
				\begin{tabular}{ccccccccccccc}
					\hline
					\multirow{2}{*}{Fault}&&  \multicolumn{2}{c}{KPCA}  &  \multicolumn{2}{c}{RPCA}  &  \textbf{RBPCA}& \multicolumn{2}{c}{DKPCA} & \textbf{DRBPCA}& \textbf{2D-RBPCA} & MV-RKPCA&\textbf{MV-RBPCA} \\
					& & $T^2$& $Q$&$T^2$ &$Q$ &$Q$ &$T^2$ &$Q$ &$Q$ & $Q$& $Q$ & $Q$\\
					\hline
					\multirow{2}{*}{No.1}&FDR&0.9975  & 0.9950 & 0.9943& 0.9977& 0.9975 & 0.9938  & 0.9913 & 0.9954 & 0.9977& 0.9875&\textbf{0.9979 } \\
					&FAR& 0.0125  & 0.0003 & 0.0112& 0.0119& 0.0106  & 0.0063  &0.0000 & 0.0170  & 0.0120 &0.0000 &0.0253\\
					\hline
					\multirow{2}{*}{No.2}&FDR&  0.9863 
					&  0.9850 
					&  0.9852&   0.9853& 0.9855& 0.9838 & 0.9825 &  0.9849 & 0.9834 &  0.9825& \textbf{0.9866}  \\
					&FAR&  0.0063 &  0.0000 &  0.0141&  0.0102&   0.0056 &  0.0188 & 0.0000 & 0.0313 & 0.0019 & 0.0063&0.0188  \\
					\hline
					\multirow{2}{*}{No.3}&FDR& 0.0350 
					& 0.0300 		
					& 0.0225& 0.0483& 0.0429 
					& 0.0163 
					& 0.0013 
					& \textbf{0.0803} 
					&0.0173 
					& 0.0000&0.0583  \\
					&FAR& 0.0000 
					& 0.0000 
					& 0.0070& 0.0269& 0.0275 & 0.0000 
					& 0.0000 
					& 0.0269 
					&0.0019 
					&  0.0000&0.0456  \\
					\hline
					\multirow{2}{*}{No.4}&FDR& \textbf{0.9999}&  0.5588 
					& 0.7116& 0.9973& 0.9960 
					& 0.0000 
					&  0.0550 
					&0.9993 
					& 0.5079 
					& 0.0000&0.9941 \\
					&FAR& 0.0063 
					& 0.0063 
					& 0.0064& 0.0117& 0.0056 
					& 0.0000 
					& 0.0000 
					&0.0325 
					& 0.0000 
					&  0.0063&0.0211\\
					\hline
					\multirow{2}{*}{No.5}&FDR&0.2825 
					& 0.3075 
					& 0.2620& 0.3564& 0.3293 
					& 0.0900 
					&0.0025 
					&0.4235 
					& 0.2573 
					&  \textbf{0.4838}&0.3820 \\
					&FAR& 0.0063 
					& 0.0063 
					& 0.0060 & 0.0060 & 0.0119 
					& 0.0000 
					& 0.0000 
					& 0.0475 
					& 0.0000 
					& 0.0063  &0.0198 \\
					\hline 
					\multirow{2}{*}{No.6}&FDR& 0.9975 
					&\textbf{0.9999} 
					& 0.9985& \textbf{0.9999} & \textbf{0.9999} 
					& 0.9938 
					& \textbf{0.9999} 
					&\textbf{0.9999} 
					&0.9983 
					&\textbf{0.9999} & \textbf{0.9999}\\
					&FAR& 0.0000 
					& 0.0000 
					& 0.0013& 0.0041& 0.0031 
					& 0.0063 
					& 0.0000 
					&0.0034 
					& 0.0000 
					&0.0000 &0.0089  \\
					\hline
					\multirow{2}{*}{No.7}&FDR& \textbf{0.9999}& \textbf{0.9999}& \textbf{0.9999}& \textbf{0.9999} & 0.9994 
					& \textbf{0.9999}
					& 0.9825 
					&\textbf{0.9999}
					& 0.9850 
					&    0.5475&\textbf{0.9999}\\
					&FAR& 0.0000 
					& 0.0000 
					& 0.0016&  0.0044& 0.0028 
					& 0.0063 
					& 0.0000 
					&0.0034 
					& 0.0003 
					& 0.0000&0.0098  \\
					\hline
					\multirow{2}{*}{No.8}&FDR& 0.9788 
					& 0.9738 
					& 0.9742& \textbf{0.9799} & 0.9793 
					& 0.9738 
					& 0.9675 
					& 0.9774 
					& 0.9721 
					& 0.9613 &\textbf{0.9799}  \\
					&FAR& 0.0000 
					& 0.0000 
					& 0.0028&  0.0108& 0.0134 
					&0.0000 
					&0.0000 
					&0.0163 
					& 0.0000 
					&  0.0000 &0.0224 \\
					\hline
					\multirow{2}{*}{No.9}&FDR& 0.0388 
					& 0.0263 
					&0.0241& 0.0440& \textbf{0.0406}& 0.0300 
					& 0.0063 
					& 0.0566 
					&0.0119 
					& 0.0000&0.0578 \\
					&FAR& 0.0375 
					& 0.0375 
					& 0.0210& 0.0703& 0.0463& 0.0500 
					&0.0000 
					& 0.0875 
					& 0.0300 
					&0.0000  &0.0878 \\
					\hline
					\multirow{2}{*}{No.10}&FDR& 0.4950 
					& 0.6850 
					& 0.3196& 0.5863& 0.5624 
					& 0.4313 
					&0.3963 
					& \textbf{0.6871} 
					&0.4735 
					& 0.0675 &0.5443 \\
					&FAR& 0.0000 
					& 0.0000 
					& 0.0039& 0.0058& 0.0066 
					& 0.0000 
					&0.0000 
					&0.0316 
					& 0.0000 
					& 0.0000 &0.0104 \\
					\hline
					\multirow{2}{*}{No.11}&FDR& 0.7675 
					& 0.5600 
					& 0.5891& 0.7195& 0.7169 
					& 0.0000 
					&0.4000 
					& \textbf{0.8948} 
					& 0.5903 
					& 0.000&0.7394 \\
					&FAR& 0.0063 
					& 0.0000 
					& 0.0048& 0.0165& 0.0141 
					& 0.0000 
					& 0.0000 
					&0.0388 
					&0.0000 
					& 0.0000&0.0260 \\
					\hline
					\multirow{2}{*}{No.12}&FDR& 0.9913 
					& 0.9888 
					& 0.9875& 0.9907& 0.9899 
					& 0.0688 
					& 0.3063 
					& \textbf{0.9983} 
					&0.9894 
					&  0.9800 &0.9909 \\
					&FAR& 0.0250 
					& 0.0000 
					& 0.0161& 0.0361& 0.0341 
					&0.0000 
					& 0.0000 
					&0.0494 
					& 0.0000 
					& 0.0187 &0.0509 \\
					\hline
					\multirow{2}{*}{No.13}&FDR& 0.9545 
					& 0.9438 
					& 0.9496 & 0.9533& 0.9523 
					& 0.0125 
					&0.5963 
					& \textbf{0.9546} 
					& 0.9431 
					& 0.9363  & 0.9518 
					\\
					&FAR& 0.0125 
					& 0.0000 
					& 0.0104 & 0.0103& 0.0103 
					&0.0000 
					& 0.0000 
					& 0.0394 
					& 0.0025 
					& 0.0000  &0.0119 
					\\
					\hline
					\multirow{2}{*}{No.14}&FDR& \textbf{0.9999}
					& 0.9988 
					& 0.9998&  0.9996&0.9996 
					& \textbf{0.9999} 
					& 0.9975 
					&0.9990 
					& 0.9996 
					&0.1700&\textbf{0.9999}\\
					&FAR& 0.0063 
					&0.0000 
					& 0.0075 & 0.0146& 0.0091 
					& 0.0125 
					& 0.0000 
					& 0.0138 
					& 0.0000 
					& 0.0000 &0.0038 
					\\
					\hline
					\multirow{2}{*}{No.15}&FDR& 0.0663 
					& 0.0613 
					&  0.0238& 0.0735& 0.0676 
					& 0.0000 
					& 0.0013 
					& \textbf{0.1476}
					&0.0588 
					&0.0100 &0.0935 
					\\
					&FAR& 0.0125 
					& 0.0000 
					& 0.0078&0.0094& 0.0092 
					& 0.0000 
					& 0.0000 
					& 0.0363 
					& 0.0000 
					&0.0000 &0.0025 
					\\
					\hline
					\multirow{2}{*}{No.16}&FDR& 0.2963 
					&  \textbf{0.6225} 
					& 0.1642& 0.5156&  0.4694 
					& 0.0013 
					&0.0013 
					& 0.4989 
					& 0.2272 
					&0.6038 &0.5695 
					\\
					&FAR& 0.0125 
					& 0.0000 
					& 0.0297& 0.1006& 0.0092 
					& 0.0000 
					&0.0000 
					& 0.0363 
					&0.0000 
					&0.0000 &0.1269 
					\\
					\hline
					\multirow{2}{*}{No.17}&FDR& 0.9588 
					&  0.8788 
					& 0.8646& 0.9562 & 0.9463 
					& 0.0000 
					& 0.5713 
					&\textbf{0.9659 }
					& 0.9429 
					&  0.3550  &0.9391 
					\\
					&FAR& 0.0000 
					& 0.0000 
					&0.0073  & 0.0107 & 0.0097 
					& 0.0000 
					& 0.0000 
					&0.0369 
					&0.0000 
					& 0.0000 &0.0050 
					\\
					\hline
					\multirow{2}{*}{No.18}&FDR&  0.9038 
					&  0.8925 
					& 0.8965 &  0.9044& 0.9029 
					&  0.8875 
					& 0.8938 
					& \textbf{0.9094}& 0.8921 
					& 0.8688  &0.9016 
					\\
					&FAR&  0.0000 
					&  0.0000 
					&0.0118 &  0.0135 &0.0134 
					&  0.0438 
					& 0.0000 
					& 0.0266 
					& 0.0000 
					&0.0000 &0.0056 
					\\
					\hline
					\multirow{2}{*}{No.19}&FDR&  0.1113 
					&  0.3813 
					&0.0981 &0.2114  &  0.1851 
					&  0.0000 
					& 0.0000 
					& \textbf{0.4899} 
					& 0.0052 & 0.0488 &0.2875 
					\\
					&FAR&  0.0063 &  0.0000& 0.0088 &  0.0133& 0.0097 &  0.0000 
					& 0.0000 & 0.0450 & 0.0000 
					& 0.0187&0.0013 
					\\
					\hline
					\multirow{2}{*}{No.20}&FDR&  0.6113 
					&  0.5688 
					& 0.4048  &0.5915  & 0.5851 
					&  0.0600 
					& 0.0013 
					& \textbf{0.7375} 
					& 0.5593 
					&   0.2275&0.6128 
					\\
					&FAR&  0.0125 
					&  0.0000 
					& 0.0069 & 0.0120 & 0.0103 
					&  0.0000 
					& 0.0000 
					& 0.0375 
					& 0.0000 
					& 0.0000&0.0056 
					\\
					\hline
					\multirow{2}{*}{No.21}&FDR&  \textbf{0.4925} 
					&  0.3813 
					& 0.4253 &  0.4611 &  0.4622 
					&  0.4375 
					& 0.2750 
					& 0.4592 
					& 0.3487 
					& 0.1288&0.4300 
					\\
					&FAR&  0.0250 
					&  0.0000 
					& 0.0203 & 0.0390 & 0.0344 
					&  0.0125 
					& 0.0000 
					& 0.0581 
					& 0.0078 
					& 0.0000 &0.0138 
					\\
					\hline
				\end{tabular}
				\label{tab:TEP}
				\begin{tablenotes}
					\item	The meaning of abbreviations is the same as in Table \ref{tab: Fault1_FDR}. The proposed methods and the most accurate values in the monitoring results for each fault have been bolded.
				\end{tablenotes}
	\end{threeparttable}}  }
\end{table*}

For processes with complex time-dependent properties, such as the Tennessee Eastman Process (see Table \ref{tab:TEP}), we observe that the process exhibits autocorrelation and does not demonstrate significant non-stationarity. Consequently, we anticipate that dynamic methods will be the most appropriate method for fault detection.
In practice, we found this to be the case to a large extent, with dynamic random Bernoulli PCA providing the most accurate fault detection among the 12 kinds of fault scenarios. Moving-window random Bernoulli PCA performs best under 6 kinds of fault scenarios compared to the time-invariant methods. The time-varying methods perform well because they cover the whole calibration phase of the process. Although various methods are practiced, none reliably detected faults 3, 9, or 15, which is consistent with the findings in \cite{Rato2016ASC}.

\begin{table*}
	\caption{The run time of the eight methods in the modeling stage and the online monitoring stage: modeling time (MT) and the average time of online monitoring (OT).}
	\renewcommand\arraystretch{1}
	{		\resizebox{\linewidth}{!}{
			\begin{threeparttable}
				\begin{tabular}{cccccccccc}
					\hline
					&&  KPCA  &  RPCA  &  \textbf{RBPCA}  & DKPCA  & \textbf{DRBPCA} & 
					\textbf{2D-RBPCA} & MV-RKPCA& \textbf{MV-RBPCA} \\
					\hline
					\multirow{2}{*}{MT(s)}&NE&  4.6391&  0.0483&  0.0485&   5.9338&   0.0475& 0.4644& 19.4380&9.7719\\
					&TEP& 5.5633 & 0.0545& 0.0538 & 6.6975 &  0.0596  & 0.4472 &10.1464& 9.2456\\
					\hline
					\multirow{2}{*}{OT(s)}& NE& 0.0051 & $2.0931\times 10^{-4}$ &  $1.2012\times 10^{-4}$& 0.0051 & $2.1601\times 10^{-4}$& 0.0015&0.3223 &$1.2690\times 10^{-4}$\\
					& TEP& 0.0050&$2.3518\times 10^{-4}$ & $1.4387\times 10^{-4}$& 0.0054&  $3.1597\times 10^{-4}$&0.0045 & 0.1164&  $1.4540\times 10^{-4}$\\
					\hline
				\end{tabular}
				\label{tab: time}
				\begin{tablenotes}
					\item The meaning of abbreviations is the same as in Table \ref{tab: Fault1_FDR}. The proposed methods have been bolded. Datasets: NE, the numerical example; TEP, the Tennessee Eastman Process.
				\end{tablenotes}
	\end{threeparttable}}}
\end{table*}

As shown in Table \ref{tab: time}, the methods utilizing random Bernoulli feature lead to a significant enhancement in computational efficiency, resulting in an order of magnitude reduction in both modeling time and online monitoring time compared to kernel-based methods. Especially for time-varying methods, the average time required to detect a single sample can be reduced by three orders of magnitude, even if the model needs to be updated several times during the online monitoring stage.

\subsection{Analysis of Individual Parameters  (m, p, l)}\label{sec: parameters}
To analyze the effects of parameters $(m,p,l)$ on the monitoring methods, only one of them is changed at a time, and the remaining parameters are kept unchanged.
Due to the limited space, here we only show the comparison results of dynamic kernel PCA, dynamic random Bernoulli PCA, and two-dimensional random Bernoulli PCA on the first datasets with Fault 1.

Firstly, the parameters $p=0.05$, $l=2$ (dynamic kernel PCA and dynamic random Bernoulli PCA), and $l=10$ (two-dimensional random Bernoulli PCA) are fixed, and the number of random Bernoulli features $m$ is increased from 50 to 500. The effect of $m$ on the performance and computational complexity is shown in Figure \ref{fig: Fault 1}.
Secondly, with fixed $m=150$, $l=2$ (dynamic kernel PCA and dynamic random Bernoulli PCA), and $l=10$ (two-dimensional random Bernoulli PCA), the effect of the parameter of  Bernoulli distribution $p$ is shown in Figure \ref{fig: Fault 1 - p}.
Finally, with fixed $m=150$ and $p=0.05$, the effect of time lag $l$ is shown in Figure \ref{fig: time lag}.
The leftmost plot compares the three methods on two performance metrics: fault detection rate and false alarm rate; The middle plot shows the computational complexity of building a model using normal operation condition samples; The rightmost plot shows the computational complexity of judging the operating condition of a single sample during online monitoring.

\begin{figure*}
	\centering
	\subfloat[accuracy: FDR and FAR]{\includegraphics[scale=0.16]{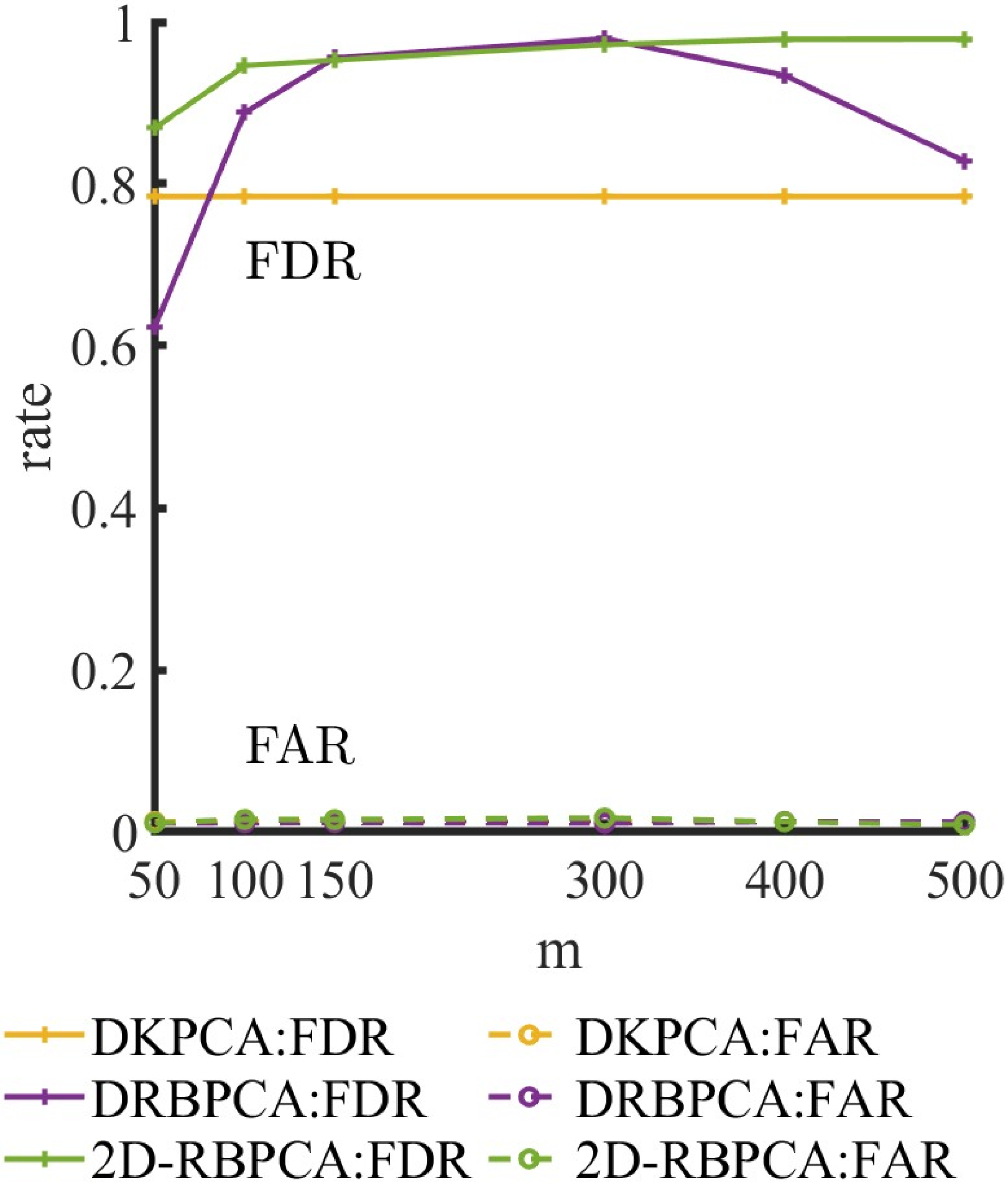}}
	\hfill
	\subfloat[complexity: modeling]{\includegraphics[scale=0.16]{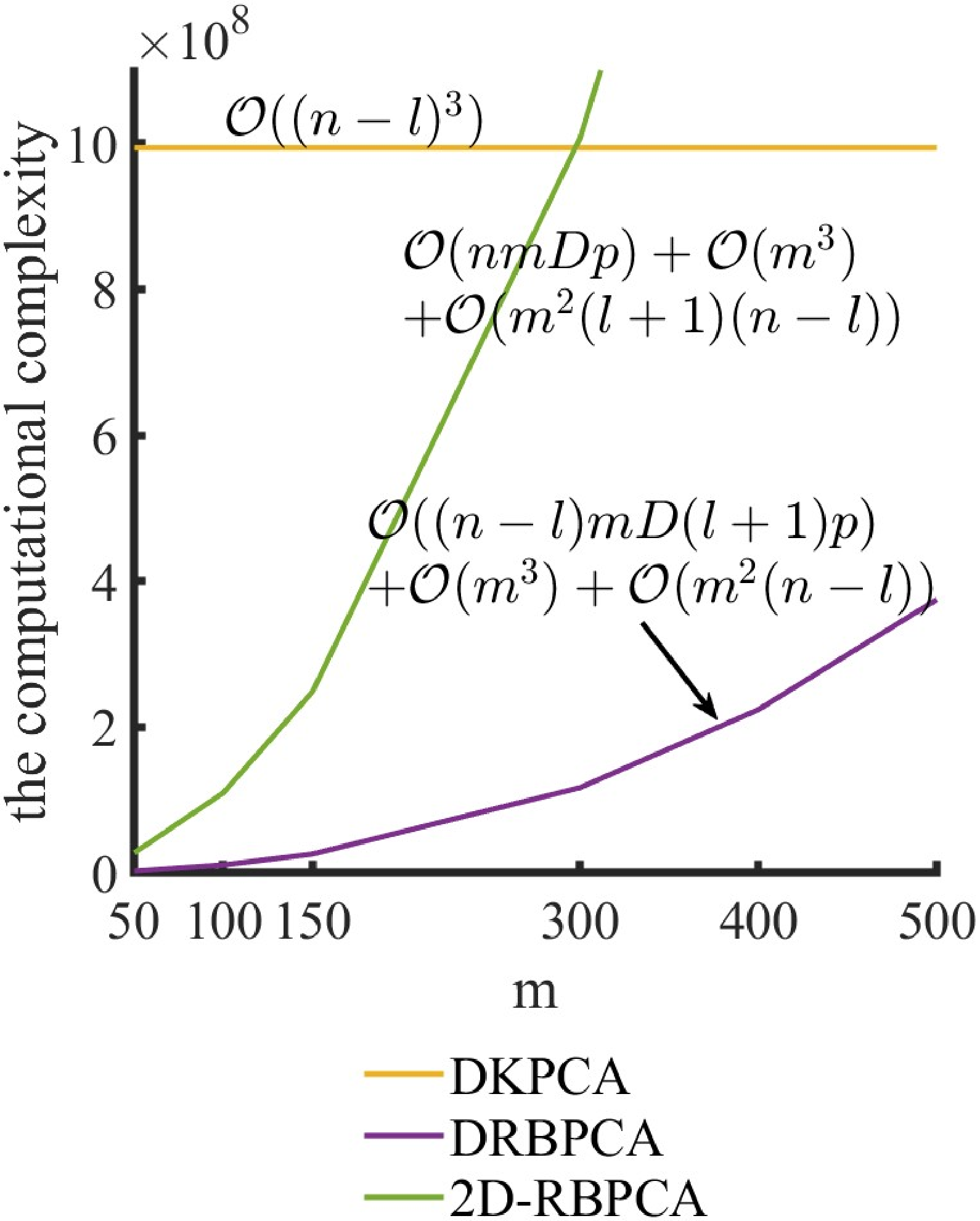}}
	\hfill
	\subfloat[complexity: online stage]{\includegraphics[scale=0.16]{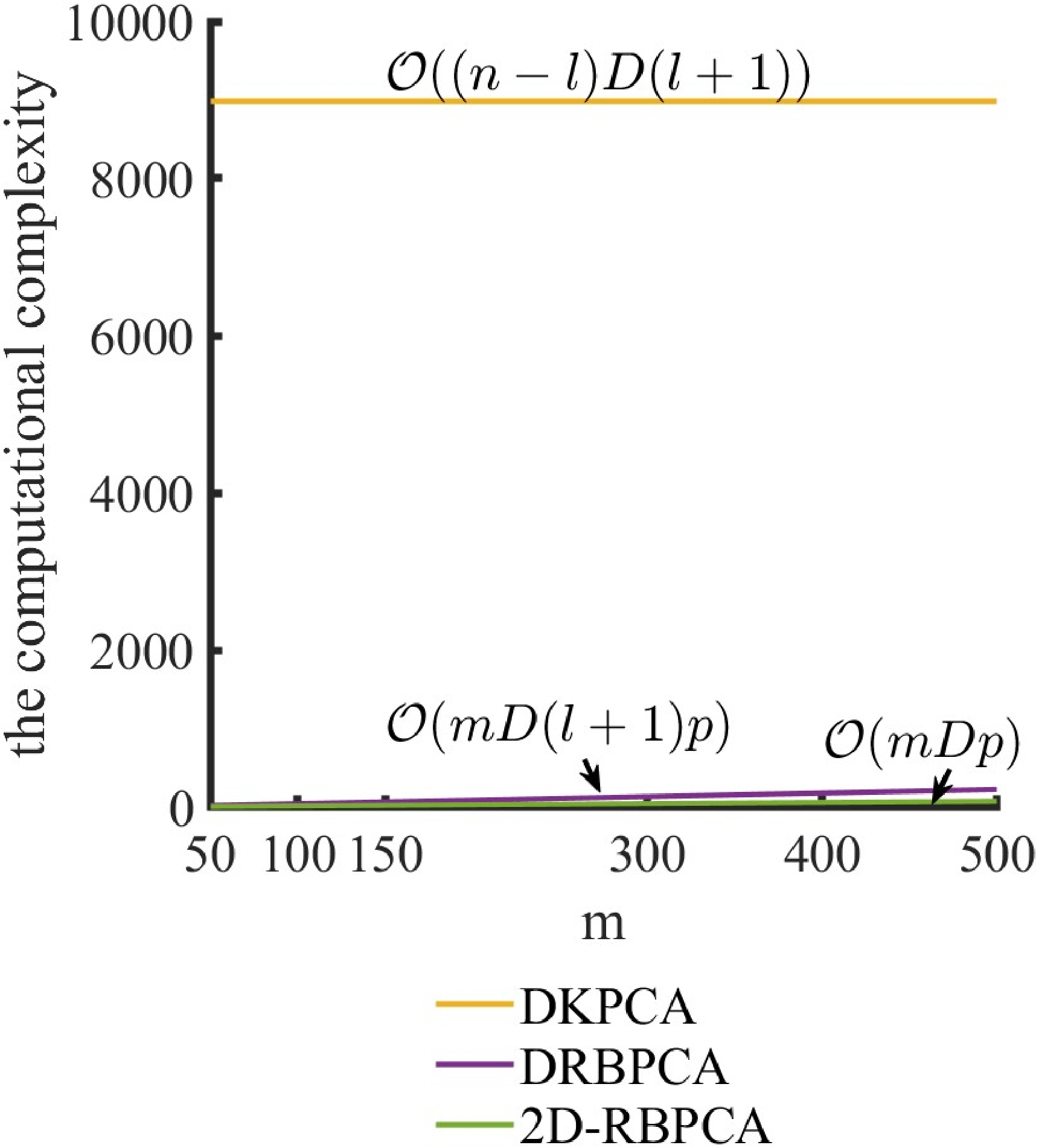}} 
	\caption{The accuracy and the computational complexity under varying numbers of random Bernoulli features $m$. FDR, fault detection rate; FAR, false alarm rate. DRBPCA, dynamic random Bernoulli PCA; 2D-RBPCA, two-dimensional random Bernoulli PCA; DKPCA, dynamic kernel PCA.}
	\label{fig: Fault 1}
\end{figure*}

Figure \ref{fig: Fault 1} shows that: (i) When $m$ is greater than 100, the detection results of the two methods, dynamic random Bernoulli PCA and two-dimensional random Bernoulli PCA,  are always better than that of dynamic kernel PCA. Moreover,  the performance of two-dimensional random Bernoulli PCA is the best and tends to be stable. (ii) In the modeling stage, dynamic random Bernoulli PCA has the lowest computational complexity, and the complexity of dynamic random Bernoulli PCA and two-dimensional random Bernoulli PCA will increase significantly with the increase of $m$, but it is still much lower than dynamic kernel PCA at $m=150$. (iii) In the online monitoring stage, the complexity of dynamic kernel PCA is much higher than that of the other two, the complexity of two-dimensional random Bernoulli PCA is slightly lower than that of dynamic random Bernoulli PCA, and the influence of $m$ on the complexity is weak.

\begin{figure*}
	\centering
	\subfloat[accuracy: FDR and FAR]{\includegraphics[scale=0.16]{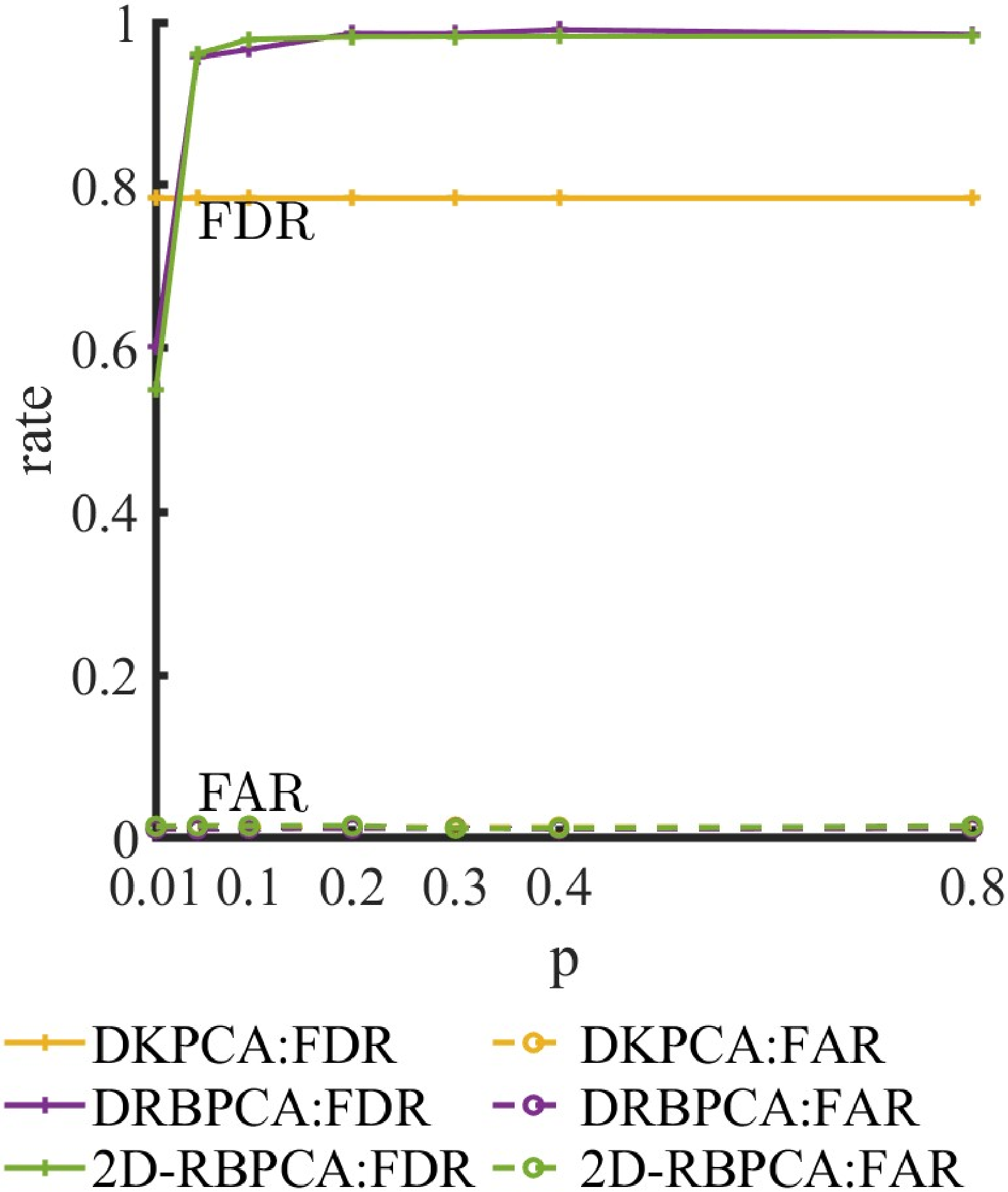}}
	\hfill
	\subfloat[complexity: modeling]{\includegraphics[scale=0.16]{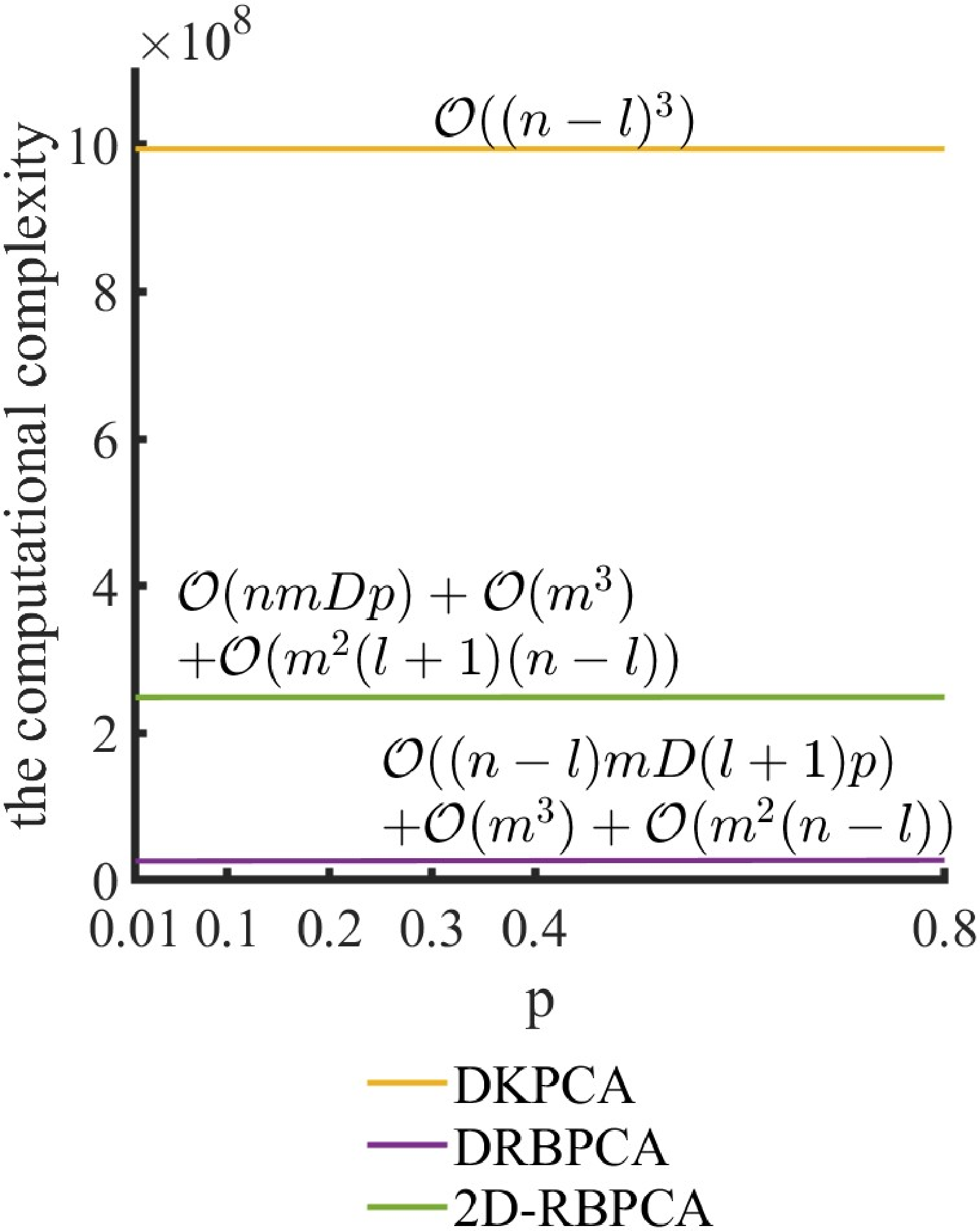}}
	\hfill
	\subfloat[complexity: online stage]{\includegraphics[scale=0.16]{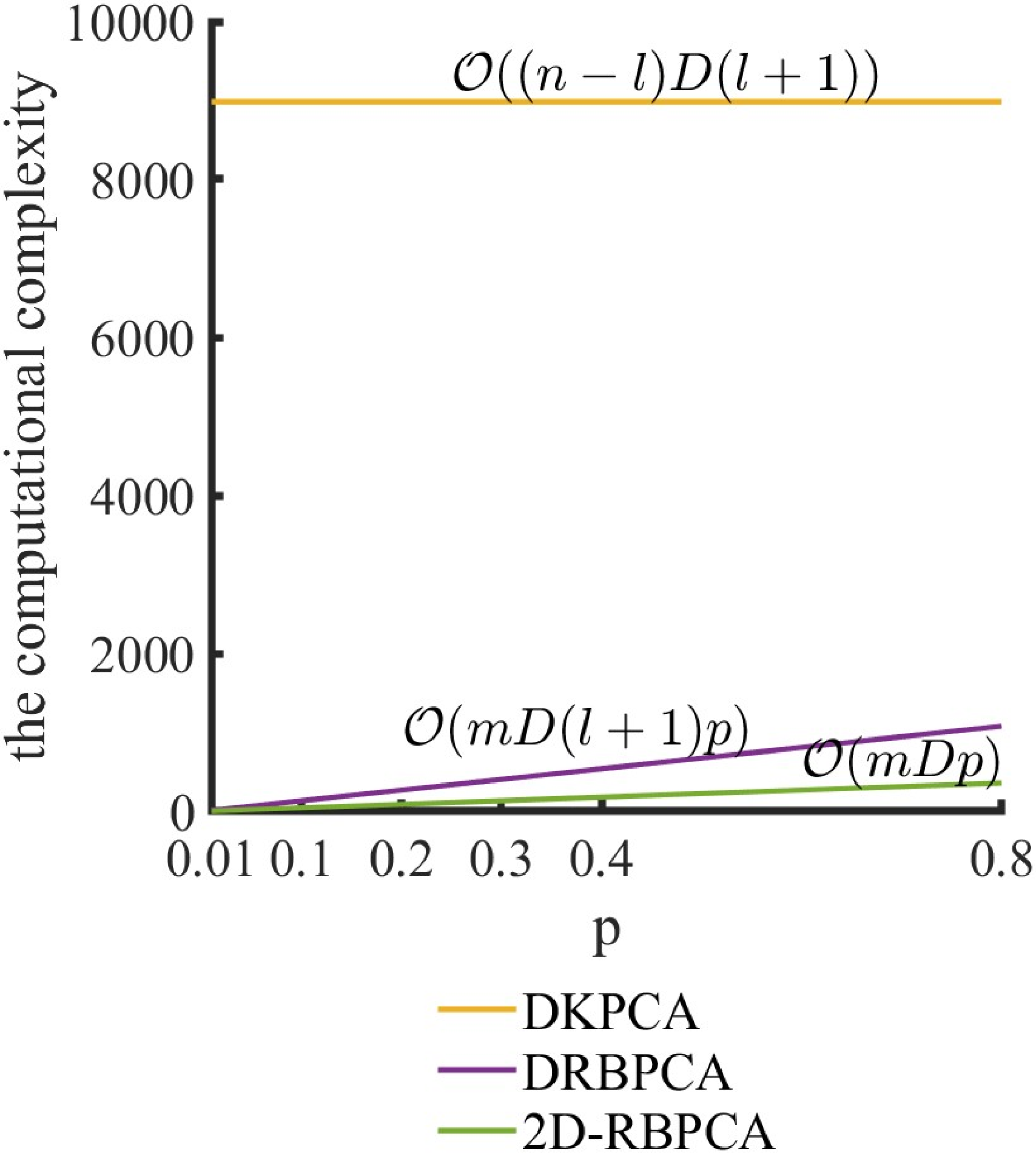}} 
	\caption{The accuracy and the computational complexity under varying parameters of Bernoulli distribution $p$. FDR, fault detection rate; FAR, false alarm rate. DRBPCA, dynamic random Bernoulli PCA; 2D-RBPCA, two-dimensional random Bernoulli PCA; DKPCA, dynamic kernel PCA.}
	\label{fig: Fault 1 - p}
\end{figure*}

Figure \ref{fig: Fault 1 - p}  illustrates that: (i) Once $p$ exceeds 0.05, both dynamic random Bernoulli PCA and two-dimensional random Bernoulli PCA exhibit higher detection rates compared to dynamic kernel PCA, and fault detection rates maintain a stable and continuous level {above} 90\%. (ii) The influence of $p$ on the computational complexity in the modeling stage is not significant, but a smaller $p$ can save more complexity when calculating the features of a single sample.

\begin{figure*}
	\centering
	\subfloat[accuracy: FDR and FAR]{\includegraphics[scale=0.16]{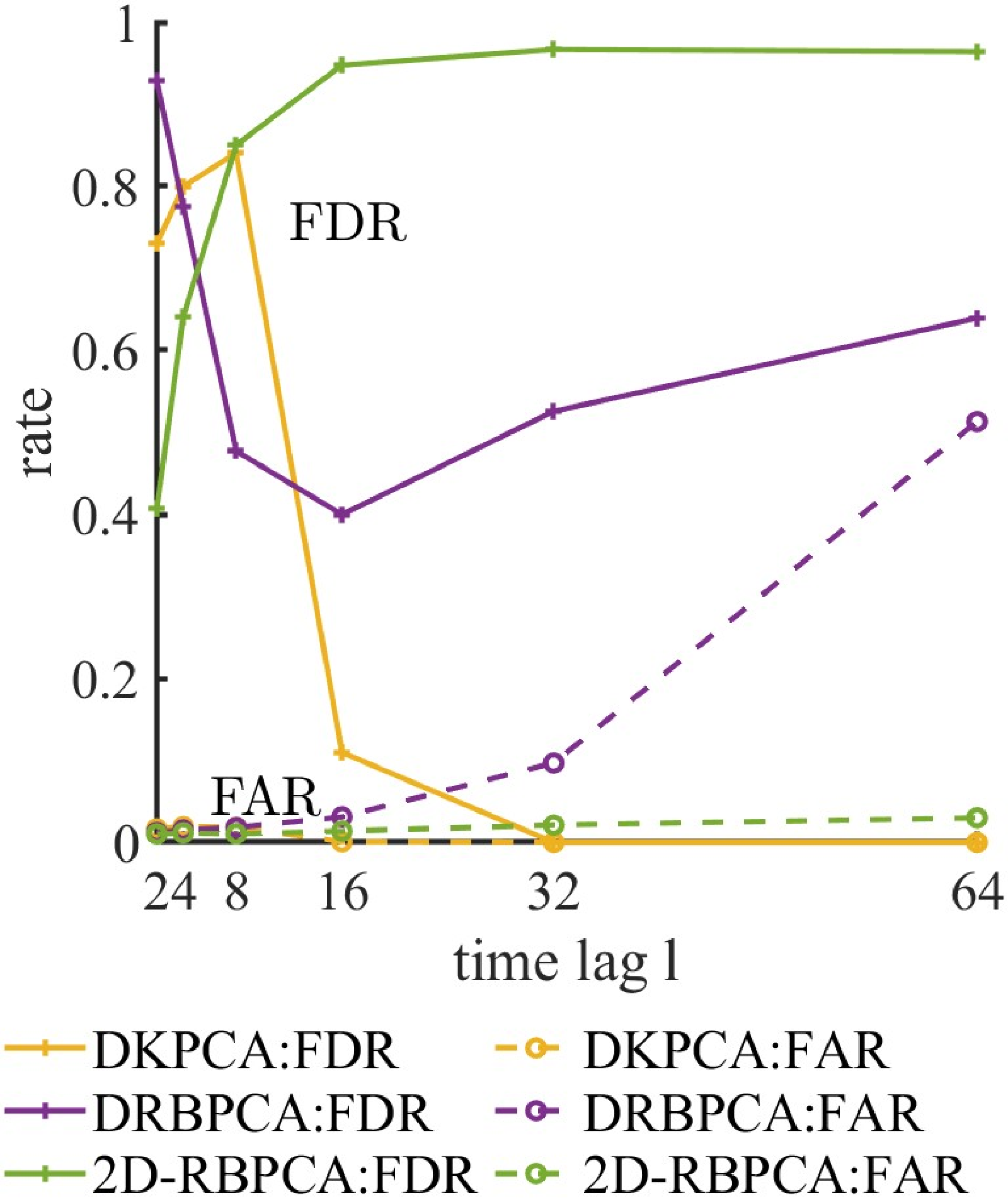}}
	\hfill
	\subfloat[complexity: modeling]{\includegraphics[scale=0.16]{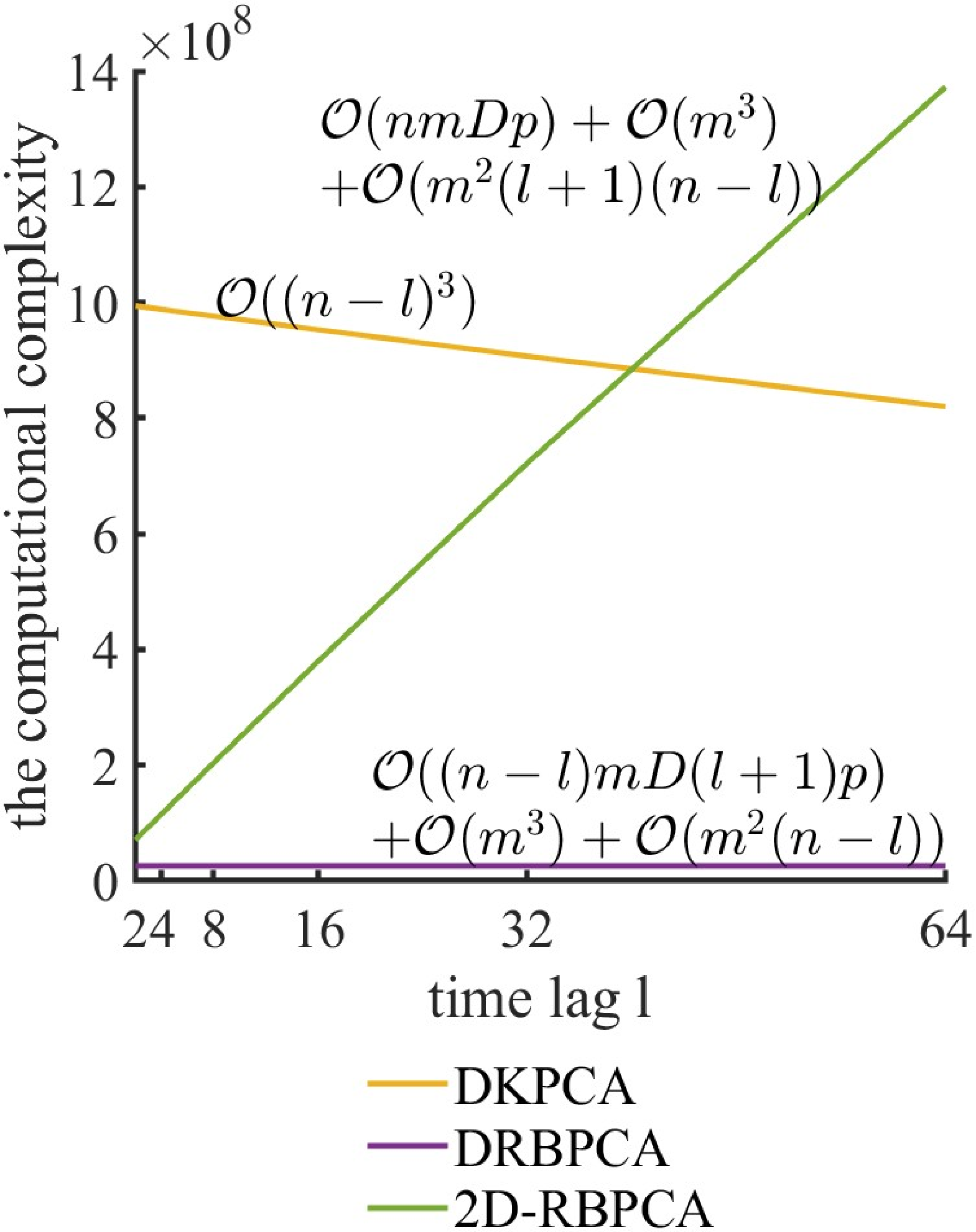}}
	\hfill
	\subfloat[complexity: online stage]{\includegraphics[scale=0.16]{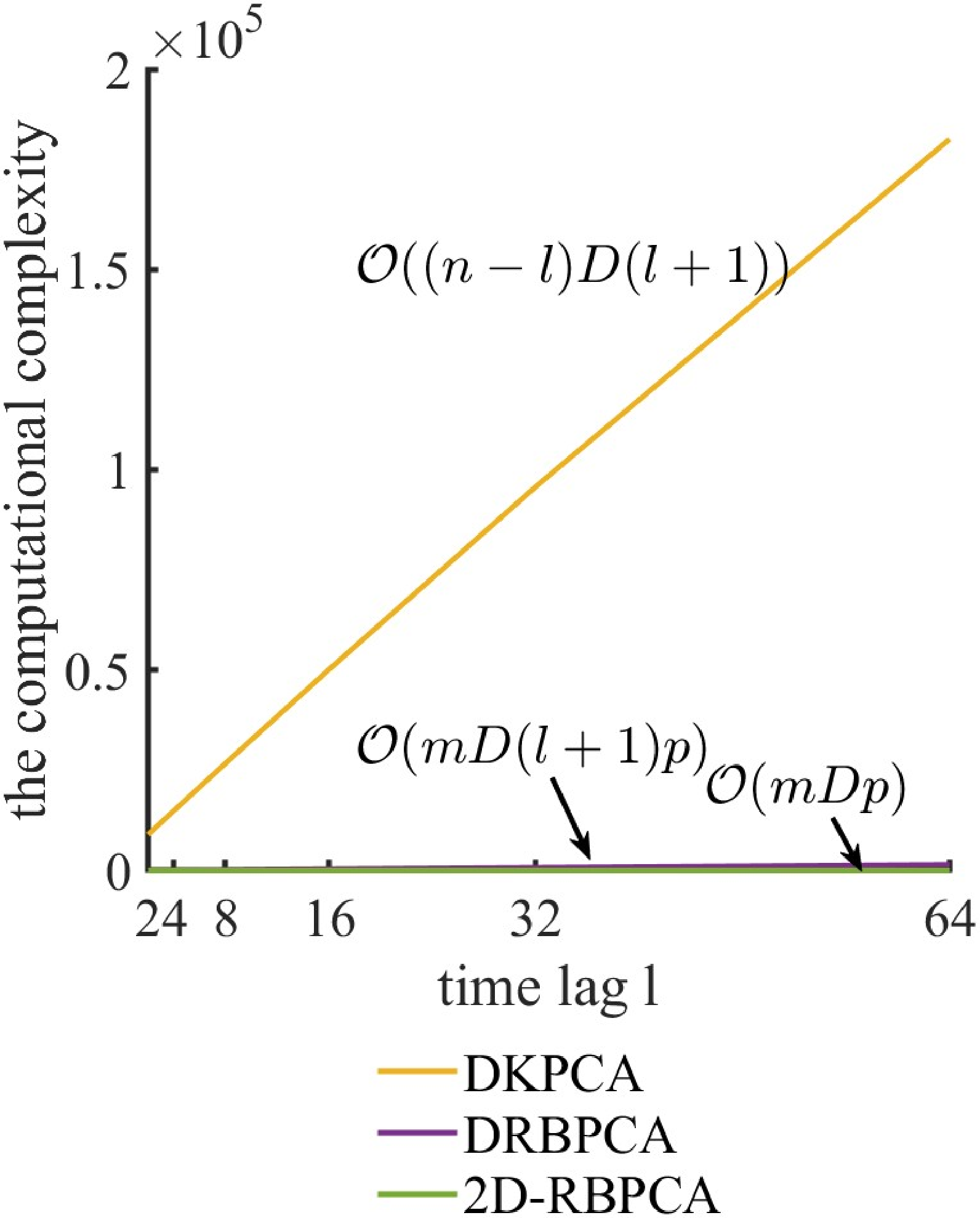}} 
	\caption{The accuracy and the computational complexity under varying time lags $l$. FDR, fault detection rate; FAR, false alarm rate. DRBPCA, dynamic random Bernoulli PCA; 2D-RBPCA, two-dimensional random Bernoulli PCA; DKPCA, dynamic kernel PCA.}
	\label{fig: time lag}
\end{figure*}

Figure \ref{fig: time lag} shows that dynamic random Bernoulli PCA and two-dimensional random Bernoulli PCA are suitable for different range of time lags. 
For small values of $l$, dynamic random Bernoulli PCA and dynamic kernel PCA behave similarly, while two-dimensional random Bernoulli PCA does not perform well. With the increase of $l$, dynamic kernel PCA directly fails, and the false alarm rate of dynamic random Bernoulli PCA rapidly increases rendering the method unusable. {Conversely,} the fault detection rate of two-dimensional random Bernoulli PCA has an obvious upward trend with $l$. Especially when $l>8$, the detection performance of two-dimensional random Bernoulli PCA far exceeds the other two methods. 
However, the complexity of two-dimensional random Bernoulli PCA increases significantly for ultra-large $l$, then the dynamic random Bernoulli PCA is recommended when rapidity is concerned with a slight loss of accuracy.

\section{Real Data: the Server Machine Dataset}
\label{Sec:Real data}
The Server Machine Dataset is a {real} dataset that \cite{10.1145/3292500.3330672} collects from a large internet company and is published publicly on \url{https://github.com/NetManAIOps/OmniAnomaly}. 
The dataset collected data from 28 machines for five consecutive weeks, with adjacent observations spaced one minute apart. We intercept part of the data from three of these machines as monitoring samples for modeling and online fault detection. Faults in the test set are marked, and more details of samples are placed in supplementary material. The monitoring performances and the run time of different methods on this dataset are shown in Table \ref{tab: SMD}.  The most accurate one for the method with two statistics is shown in the table.

\begin{table*}[ht]
	\caption{The comparative fault detection rate (FDR), false alarm rate (FAR), the modeling time (MT), and the average online monitoring time (OT) on the Server Machine Dataset.}
	\renewcommand\arraystretch{1}
	{		\resizebox{\linewidth}{!}{
			\begin{threeparttable}
				\begin{tabular}{cccccccccc}
					\hline
					Sample& & KPCA  &  RPCA  &  \textbf{RBPCA}  & DKPCA  & \textbf{DRBPCA} & 
					\textbf{2D-RBPCA} & MV-RKPCA& \textbf{MV-RBPCA} \\
					\hline
					\multirow{4}{*}{No.1}&FDR&0.8970  & 0.8917 & 0.8169 &  0.7774  & \textbf{0.9161}  & 0.7859& 0.8967& 0.8831 \\
					&FAR& 0.0283 & 0.0181 & 0.0185 & 0.0167 & 0.0416   & 0.0295 &0.0483  &0.0130  \\
					& MT(s)&  1.5312
					&0.0621 & 0.0500& 4.9122&0.0605 & 0.3907&6.6863 &17.6965\\
					& OT(s)& 0.0040&$7.4832\times10^{-5}$ & $7.5440\times10^{-5}$ &  0.0043 &$1.2671\times10^{-4}$ &0.0040 &0.0974 &$5.4334\times10^{-4}$\\
					\hline
					\multirow{4}{*}{No.2}& FDR& 0.7339 & 0.6995 &0.6969  &0.1579  & \textbf{0.8856}  &0.7394  & 0.7222&0.7006 \\
					& FAR& 0.0146&0.0178 &0.0166 & 0.0223&0.0250 &0.0214 & 0.0162&0.0474 \\
					& MT(s)& 17.6527& 0.3647& 0.3414& 22.3728& 0.4865& 14.0526& 57.8884&50.9773\\
					& OT(s)& 0.0185&$3.5892\times10^{-4}$ & $3.6722\times10^{-4}$& 0.0192& $5.4539\times10^{-4}$&0.1023 &  0.5978&0.0017\\
					\hline
					\multirow{4}{*}{No.3}& FDR& 0.7034 & 0.6322 & 0.7420 & 0.7402 &  0.7709 & \textbf{0.8273} & 0.5729& 0.6723 \\
					& FAR&0.0240  &0.0481 &  0.0374 &0.0253 &0.0421 & 0.0338&0.0122 &0.0251 \\
					& MT(s)&45.1083 &0.3238 &0.3206 & 17.4680& 0.3323& 1.6163&61.8315 &51.0502\\
					& OT(s)&0.0199 &$3.4900\times10^{-4}$ &$3.5916\times10^{-4}$ &  0.0196&$3.8770\times10^{-4}$ & 0.0292& 0.7422 &0.0019\\
					\hline
				\end{tabular}
				\label{tab: SMD}
				\begin{tablenotes}
					\item The proposed methods: RBPCA, random Bernoulli PCA; DRBPCA, dynamic random Bernoulli PCA; 2D-RBPCA, two-dimensional random Bernoulli PCA; MV-RBPCA, moving-window random Bernoulli PCA, all of which have been bold.
					Other methods: KPCA, kernel PCA; RPCA, random PCA; DKPCA, dynamic kernel PCA; MV-RKPCA, moving-window reduced kernel PCA. The most accurate values 
					for each sample have been bolded.
				\end{tablenotes}
	\end{threeparttable}} }
\end{table*}

The two dynamic methods based on random Bernoulli PCA have the best performance and far outperform the kernel-based methods in the No.2 and No.3 samples. Other methods, except dynamic kernel PCA, also have a similar level of performance. 
The dynamic kernel PCA is significantly influenced by the chosen relatively large time lag. 
The detection rate of the time-varying methods with a moving window is only slightly higher or a little different from that of the static methods. Due to the complexity of real data, the screened dataset may lose information and the update criterion cannot comprehensively cover the new information in the system.

The computational speed of random Bernoulli PCA is much faster than kernel PCA in both the modeling stage and the online monitoring stage. This is because random Bernoulli features can be combined with linear algorithms to solve nonlinear problems, thus reducing the linear complexity in the sample size. Therefore, the subsequent dynamic method, dynamic random Bernoulli PCA, is at least two orders of magnitude faster than the kernel-based method, dynamic kernel PCA, in terms of computation time. The two-dimensional random Bernoulli PCA based on the time-lagged matrix is also a little faster than the dynamic kernel PCA. The time-varying method, moving-window random Bernoulli PCA, takes time to calculate the similarity when screening the data in the modeling stage, but the speed of monitoring single data is still much faster than moving-window reduced kernel PCA.

\section{Conclusion}
\label{sec:conc}
This paper proposes four fast methods based on random Bernoulli PCA for fault detection in process monitoring with different  fault scenarios.
The random Bernoulli feature integrates the bootstrap resampling into random feature mapping, significantly accelerating large-scale dense matrix multiplication and reducing computational complexity by 
$(1-p)$\%, where $p$ can be set to a very small  probability. 
The proposed random Bernoulli PCA not only retains the advantages of kernel PCA over other nonlinear PCA techniques, but also mitigates the excessive computational burden associated with the dimensionality of kernel matrix being equal to the number of samples in the input space.
The experimental results show that monitoring methods based on random Bernoulli features can achieve comparable performance to kernel-based ones in most cases and sometimes even outperform it, but with much lower computational cost.
Both modeling and online monitoring times are reduced by at least one order of magnitude, since only the linear algorithm is needed in the mapped feature space.
Consequently, the application of random Bernoulli features can be further extended  as a more efficient solution to other nonlinear problems.

\section*{Acknowledgement}
The authors are grateful to the Editor, the Associate Editors, and the referees for their review of the paper.
The authors are supported by Key technologies for coordination and interoperation of power distribution service
resource, Grant No. 2021YFB2401300.

\section*{Supplementary Materials}
The supplementary material includes the analysis of the Gaussian kernel estimator based on random Bernoulli feature, the proof of lemma and theorem in Section \ref{Sec:Random Bernoulli PCA}, the algorithms of dynamic and time-varying process monitoring, the comparison of spectral properties of approximate kernel matrices, and more details on real data and parameter settings. The code is made available at \url{https://github.com/kchen-2024/RBPCA.git}.

\bibliographystyle{elsarticle-num}      
\bibliography{template-bib}  
\end{document}